# Optimizing the Neural Network Training for OCR Error Correction of Historical Hebrew Texts

Omri Suissa[1], Avshalom Elmalech, and Maayan Zhitomirsky-Geffet

[1] Bar Ilan University, Department of Information Science, Ramat Gan 52900, Israel
omrivm@gmail.com

**Abstract.** Over the past few decades, large archives of paper-based documents such as books and newspapers have been digitized using Optical Character Recognition. This technology is error-prone, especially for historical documents. To correct OCR errors, post-processing algorithms have been proposed based on natural language analysis and machine learning techniques such as neural networks. Neural network's disadvantage is the vast amount of manually labeled data required for training, which is often unavailable. This paper proposes an innovative method for training a light-weight neural network for Hebrew OCR post-correction using significantly less manually created data. The main research goal is to develop a method for automatically generating language and task-specific training data to improve the neural network results for OCR post-correction, and to investigate which type of dataset is the most effective for OCR post-correction of historical documents. To this end, a series of experiments using several datasets was conducted. The evaluation corpus was based on Hebrew newspapers from the JPress project. An analysis of historical OCRed newspapers was done to learn common language and corpus-specific OCR errors. We found that training the network using the proposed method is more effective than using randomly generated errors. The results also show that the performance of the neural network for OCR post-correction strongly depends on the genre and area of the training data. Moreover, neural networks that were trained with the proposed method outperform other state-of-the-art neural networks for OCR post-correction and complex spellcheckers. These results may have practical implications for many digital humanities projects.

**Keywords:** OCR Post-correction, Neural Networks, Hebrew Historical Newspapers, Digital Humanities.

## 1 Introduction

Over the last few decades, massive digitization of historical document collections has been performed using OCR techniques. As a result, large digital repositories have been created, e.g., the Library of Congress's historical digital collection [20[ and the British Newspaper Archive [15] with various discovery tools (e.g., [6]). Even commercial enterprises have initiated large-scale OCR projects like Google Books [16].

An OCR algorithm processes a high-resolution image of the resource (e.g., a book or newspaper page) and converts it into text. Unfortunately, OCR output for historical documents is often inaccurate. OCR errors, sometimes called spelling mistakes, come in several forms: insertions, deletions, substitutions, transposition of characters, splitting and combining of words [11].

Digitization is essential for preservation and increasing the accessibility and research of cultural heritage. Thus, in many digital humanities projects which use digitized historical collections, there is a need to search and automatically analyze the text of the documents. However, OCR errors undermine the research and preservation efforts. Therefore, improving the quality of the OCR technology has recently become a critical task. Numerous studies applied machine learning techniques to correct OCR errors [1, 10]. One of the most effective machine learning approaches is deep learning based on multi-layer neural networks, which have been successfully applied in many document processing tasks, including the spellchecking for modern texts [10]. However, the utilization of neural networks for OCR error correction in historical documents is still underexplored in previous research [1]. Particularly, there is no available effective neural network model for fixing OCR errors in historical Hebrew newspapers. Hence, the primary goal of this research is to develop an effective methodology for designing an optimized neural network for OCR post-correction for Hebrew historical texts with a minimal amount of manually created training data.

Neural networks are a componential model built using "neurons" in "layers". Each neuron gets an input, performs a mathematical calculation, and transfers its result (output) to other neurons. The first layer receives the task's input, which is transferred through the network, and the last layer's output is the predicted result of the network [13, 14]. The main advantage of neural networks is their ability to automatically calculate the optimal representative feature set for the given task rather than relying on manually selected features. As a baseline of the study, we used the neural network model from Ghosh & Kristensson [3] that was designed for OCR post-correction. This network was based on the Gated Recurrent Units (GRU) [2] architecture. We also tested the Long Short Term Memory (LSTM) [5] architecture, which was found effective in various NLP (natural language processing) tasks [9].

To build an optimal model for a specified task, a neural network has to be trained on a certain dataset for which both the input (OCRed text) and the target data (correct golden standard text) are provided. In this study, we investigated the influence of the training dataset characteristics on the network's performance. In particular, we experimented with different types of training datasets from various genres (secular literature vs. the Bible) and historical periods (from the last two centuries, ancient and modern), as well as with different types of OCR errors (random OCR errors vs. language and corpus specific OCR errors). Finally, we compared and analyzed the accuracy of the obtained networks in OCR error correction of Hebrew historical newspapers from the JPress corpus [21].

## 2 Methodology

### 2.1 Dataset Generation

The evaluation dataset of the study (JP_CE) was created from 150 OCRed historical Hebrew newspapers articles randomly selected from JPress - the most extensive historical Hebrew newspapers collection, dated 1800-2015 [17]. The articles included OCR errors, which were manually fixed by 75 students. The students' corrections were double-checked by an expert to create a high-quality golden standard corpus. This dataset comprised of the original and corrected versions of the above 150 JPress articles was used to evaluate the networks' performance.

Next, four different training datasets were generated as follows. Each dataset comprised two versions of the same texts – the artificially created OCRed text and its golden standard version. Two datasets were based on texts from the Ben Yehuda Project [18] (the Hebrew equivalent of the Gutenberg [19] project comprised of secular Hebrew literature mostly from the last two centuries and the Middle ages), and two others consisted of the Hebrew Bible text. Both the Ben-Yehuda and Bible texts were typed manually and are thus considered correct. Each of them belongs to a different time period and genre, while Ben-Yehuda's period (partially) overlaps with that of the JPress corpus. To create training sets with OCR errors, we intentionally inserted errors in each of the above corpora (Ben-Yehuda and the Bible) using two different methods. The first one was a random error generation procedure [11,4], when randomly chosen characters in each line of the text were removed, replaced (with other randomly chosen characters), or inserted at a randomly selected position. As a result, BYP and BIBLE datasets were created (as shown in Table 1). The alternative approach was to insert language and corpus-specific OCR errors, automatically learned from the JPress newspaper collection, in addition to the random error generation. The pseudo-code of the error generation algorithm is displayed in Figure 1. As can be observed from Figure 1, first, the algorithm generates some language and corpus independent types of errors, such as the removal and insertion of characters and swapping between two consecutive characters at random positions. Next, the most common JPress-specific OCR errors are added according to their relative frequency of occurrence in the corpus. To learn the most common character confusion pairs, 70% of the original JP_CE corpus and its fixed golden standard version were compared using the Needleman–Wunsch alignment algorithm [8]. The most common OCR confusion errors, along with their frequencies in JP_CE, are shown in Table 2. The outcome of this method was the BYP-HEB and BIBLE-HEB datasets.

**Table 1.** The study's datasets

| Dataset Name | Input Corpus | Target Golden Standard Corpus | Generation method |
|---|---|---|---|
| JP_CE | JPress – OCRed historical newspapers | Fixed JPress articles | Manually fixed |
| BYP | The Ben Yehuda Project with random OCR errors | The Ben Yehuda Project - books | Automatically inserted errors |
| BYP_HEB | The Ben Yehuda Project Hebrew JPress specific OCR errors | The Ben Yehuda Project - books | Automatically inserted errors |
| BIBLE | The Bible with Random OCR errors | The Hebrew Bible from sefaria.org.il | Automatically inserted errors |
| BIBLE_HEB | The Bible with Hebrew JPress specific OCR errors | The Hebrew Bible | Automatically inserted errors |

**Table 2.** Common OCR errors in Hebrew historical newspapers in JPress

| Character | Fix | Frequency |
|---|---|---|
| ח | ה | 499 |
| ד | ר | 306 |
| ג | נ | 256 |
| ב | כ | 210 |
| ׳ | , | 207 |
| ם | ס | 194 |
| ח | ת | 162 |
| ו | ׳ | 162 |

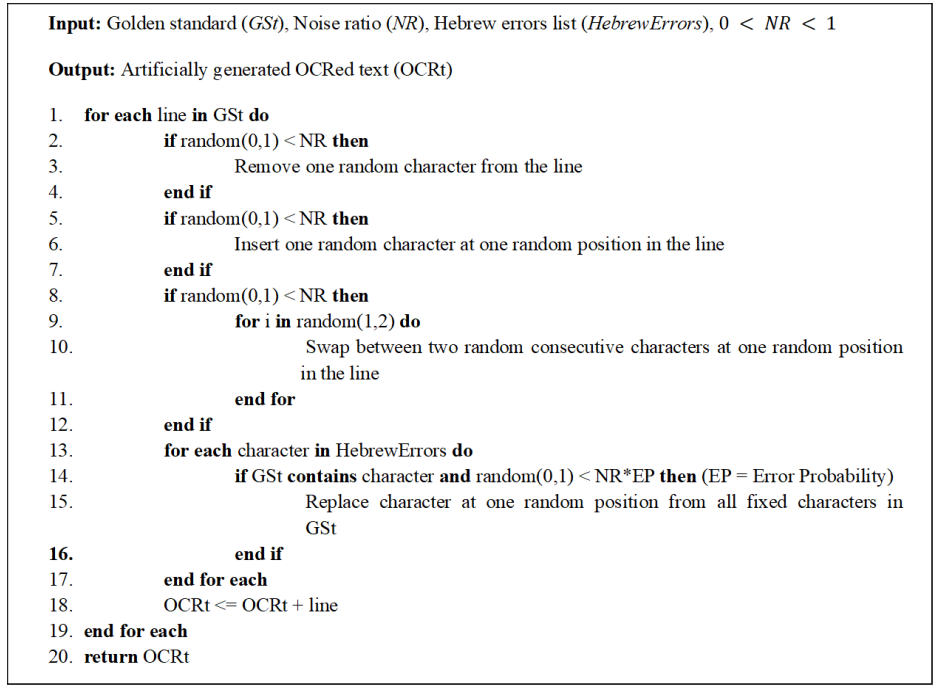

**Fig. 1.** Random and JPress-specific OCR error generation algorithm.

Figure 2 summarizes the proposed approach for constructing the neural network for OCR error correction.

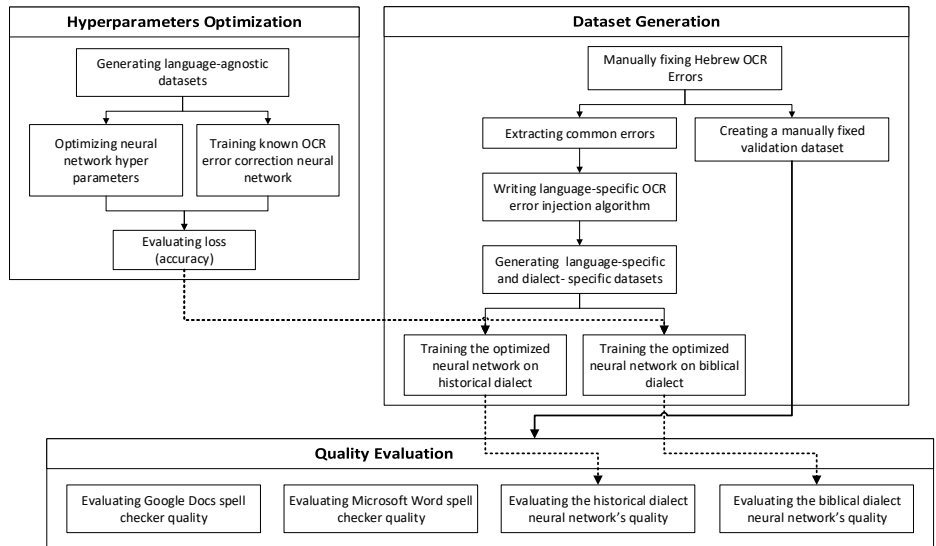

**Fig. 2.** The study's methodology diagram.

## 2.2 Evaluation Measures

To assess the quality of the results, two evaluation measures were used: 1) the character-based accuracy increase, and 2) the word-based overall accuracy of the text. The character-based increase in the text's accuracy is computed as a percentage of the errors fixed by the network out of the total number of OCR errors in the input text. The number of network's corrections is calculated as a difference between the Levenshtein's minimal edit distance [7], denoted as lev, of the input OCRed text from the )correct) golden standard version of the text, GS, and the minimal edit distance of the fixed text, Fixed, (after the network's corrections) from the golden standard text. The initial number of errors in the OCRed text is computed as the minimal edit distance between the OCRed text and the golden standard text. If a network has inserted more errors than it has fixed, the accuracy increase value is set to 0. More formally, we define acc-increase as follows:

$$acc - increase = \begin{cases} \frac{lev_{GS,OCRed} - lev_{GS,Fixed}}{lev_{GS,OCRed}} * 100, & lev_{GS,OCRed} \geq lev_{GS,Fixed} \\ 0, & otherwise. \end{cases} \quad (1)$$

To estimate the accuracy of the given text at the word-level, Wunsch alignment algorithm [8] was applied to compare the evaluated text with its golden standard version. Then, the output of the alignment was processed to split the text into words using a standard set of delimiters. The word-based accuracy of the text compared to its golden standard version is assessed with the standard word accuracy measure [22]:

$$WAcc = \frac{N_w - I_w \pm S_w \pm D_w}{N_w} * 100 \quad (2)$$

where $N_w$ is the total number of words in the evaluated text, $S_w$ is the number of words in the evaluated text that are substituted with other words in the golden standard version of the text, $D_w$ is the number of words in the evaluated text that are absent from the golden standard text, and $I_w$ is the number of words which occur in the golden standard text, but are absent from the evaluated text. The word-based metric is crucial from the user perspective since users comprehend and search texts by whole words.

## 3 Results

First, to select the most effective network model for the task, we comparatively evaluated the performance of the baseline GRU network [3] and an LSTM-based model with different hyperparameters. The optimized network was the bidirectional LSTM [12] with 4 layers, a dropout of 0.2, 500 units, an epoch size of 250,000, and a batch size of 256. The technical details of the network optimization procedure are beyond the scope of the paper.

### 3.1 The Networks' Training and Validation

To train and validate the networks, we divided each of the two datasets (BYP and BYP_HEB) described above into training (80%) and validation (20%) subsets. Then, two different networks were constructed and trained on the training subsets. The results of the networks' validation on the corresponding validation sets are presented in Figure 3. As can be observed from Figure 3, the network that was trained and validated on the BYP_HEB dataset achieved higher accuracy (94%) than the network trained and validated on BYP (85%). We concluded that training on the dataset with JPress-specific errors is more effective than training on the dataset with randomly generated errors.

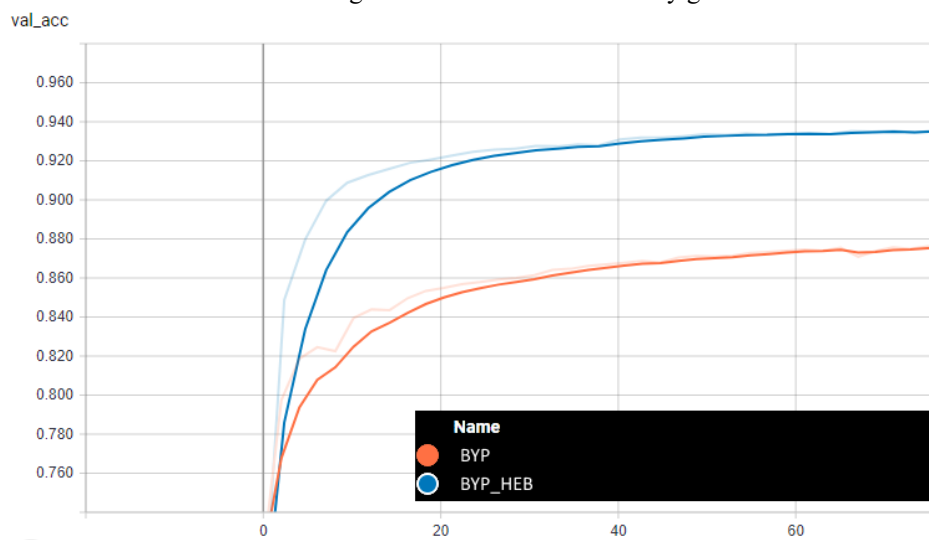

**Fig. 3.** BYP and BYP_HEB validation accuracy

### 3.2 The Networks' Evaluation

The networks' evaluation was performed by applying the two best networks (trained on BYP-HEB and BIBLE-HEB) to fix the JP_CE (historical newspapers from JPress) dataset. Note that the baseline word-based accuracy of the original evaluation dataset (JP_CE) was 48.984% (i.e., only about 49% of the words were correct before applying the networks).

In addition to the two networks trained on historical texts (Ben-Yehuda and the Bible), we evaluated the performance of the state-of-the-art spellcheckers that were implemented by Google and Microsoft as deep neural networks, trained mostly on modern Hebrew texts. Interestingly, neither Google Docs nor Microsoft Word 2019 improved the text's accuracy. Their quality score was about 0%, since they have introduced as many errors as they have fixed. From an examination of 20% of randomly chosen texts, it seems that these spellcheckers fixed well non-real words, but failed on real words (that do not make sense in the context of the sentence). Non-real words always got a fix, but not always a correct one. The spellcheckers were able to fix the following error types:

- Characters' transposition
- Redundant spacing
- "Dirt" signs (smudges, actual dirt, damaged paper)
- Real word spelling mistakes

The evaluation results are presented in Table 3. The obtained results show the dependency of the network's effectiveness on the time period of the training dataset. When the network learns from the corpus written in a similar period, it achieves positive and much better results (around 4.5% character-based and 5.5% word-based accuracy increase), than networks trained on texts from substantially more distant periods (which demonstrated none or negative change in the accuracy).

The best network (BYP_HEB) learned different types of corrections and successfully applied them on historical newspapers, including:

- Fixing spelling mistakes
- Fixing characters transposition
- Removing redundant spacing
- Adding spacing
- Preserving the names of the entities
- Removing "dirt" signs

However, the majority of the errors were not fixed by the network, and in some cases, it even introduced new errors. This might be explained by genre and style-driven differences among the training (Ben-Yehuda corpus, literature) and the evaluation datasets (JP_CE, newspaper articles).

Table 3. Comparison of all the networks evaluated on JP_CE

| Network | Character-based Accuracy Increase | Word Accuracy |
|---|---|---|
| Neural Network (BYP_HEB) | %5.406 | %53.472 |
| Google Docs spell checker | ~ 0% | %41.58 |
| Microsoft Word spell checker | ~ 0% | %41.53 |
| Neural Network (BIBLE_HEB) | ~ 0% | ~ 0% |

## 4      Conclusions

This work introduced a light-weight method to train neural networks for Hebrew OCR error post-correction. As demonstrated in the results section, there is a substantial benefit for generating a language and period-specific dataset for OCR post-correction. Interestingly, generating only a language-specific dataset using the Bible introduces more errors than corrections. It is similar to a time traveler from the biblical era trying to fix OCR errors of more modern texts.

In addition, only 105 manually fixed articles were needed for the error generation algorithm for Hebrew historical newspapers, which is a minimal human effort compared to the vast amount of labeled training data typically required for a neural network.

These results are another step towards creating automated error correction of historical Hebrew OCRed documents and historical-cultural preservation in general. Although the scope of this research was Hebrew, we believe the proposed methodology can be generalized to other languages. Researchers can use these results to reduce the complexity when designing neural networks for OCR post-correction and to improve the OCRed document correction process for many digital humanities projects.

## References

1. Chiron, G., Doucet, A., & Moreux, J.: Competition on Post-OCR Text Correction. In: ICDAR2017 Competition on Post-OCR Text Correction, pp. 1423–1428. (2017).
2. Cho, K., van Merrienboer, B., Gulcehre, C., Bahdanau, D., Bougares, F., Schwenk, H., & Bengio, Y.: Learning Phrase Representations using RNN Encoder-Decoder for Statistical Machine Translation. In: Proceedings of the 2014 Conference on Empirical Methods in Natural Language Processing (EMNLP) (2014).
3. Ghosh, S., & Kristensson, P. O.: Neural Networks for Text Correction and Completion in Keyboard Decoding. (2017).
4. D'hondt, E., Grouin, C., & Grau, B.: Generating a Training Corpus for OCR Post-Correction Using Encoder-Decoder Model. In: Proceedings of the Eighth International Joint Conference on Natural Language Processing. vol. 1, pp. 1006-1014. (2017).
5. Hochreiter, S., & Schmidhuber, J. J.: Long short-term memory. Neural Computation, 9(8), 1–32 (1997).
6. Lansdall-Welfare, T., Sudhahar, S., Thompson, J., Lewis, J., & Cristianini, N.: Content analysis of 150 years of British periodicals. In: Proceedings of the National Academy of Sciences. vol. 114 (4), pp. 457–465. (2017)
7. Levenshtein, Vladimir I.: Binary codes capable of correcting deletions, insertions, and reversals. Soviet physics doklady 10(8), 707-710 (1966).
8. Needle, S. B., Christus, A. S. D., & Needleman, Saul B., and C. D. W.: A General Method Applicable to the Search for Similarities in the Amino Acid Sequence of Two Proteins. Journal of Molecular Biology, 48(3), 443–453 (1970).
9. Pascanu, R., Mikolov, T., & Bengio, Y.: On the difficulty of training Recurrent Neural Networks. In: International conference on machine learning, pp. 1310-1318. (2013).
10. Raaijmakers, S.: A Deep Graphical Model for Spelling Correction. In: BNAIC 2013, Proceedings of the 25th Benelux Conference on Artificial Intelligence. Delft, The Netherlands (2013).

11. Reynaert, M.: Non-interactive OCR post-correction for giga-scale digitization projects. In: Lecture Notes in Computer Science, LNCS, vol. 4919, pp. 617–630. Springer, Heidelberg (2008).
12. Schuster, M., & Paliwal, K. K.: Bidirectional recurrent neural networks. IEEE Transactions on Signal Processing, 45(11), 2673–2681 (1997).
13. Sutskever, I., Vinyals, O., & Le, Q. V.: Sequence to Sequence Learning with Neural Networks. In Advances in neural information processing systems, 3104-3112 (2014)
14. Werbos, P.: Beyond Regression: New Tools for Prediction and Analysis in the Behavioral Sciences. Ph. D. dissertation, Harvard University. (1974).
15. The British Newspaper Archive, https://www.britishnewspaperarchive.co.uk/, last accessed 2019/06/20.
16. NYTimes.com, "Google Books: A Complex and Controversial Experiment", last accessed 2015/10/28.
17. JPress Collection Homepage, http://web.nli.org.il/sites/JPress/Hebrew/Pages/default.aspx, last accessed 2019/06/20.
18. Project Ben Yehuda Homepage, https://bybe.benyehuda.org/, last accessed 2019/06/20.
19. Project Gutenberg Homepage, https://www.gutenberg.org/, last accessed 2019/06/20.
20. https://chroniclingamerica.loc.gov/
21. http://www.jpress.nli.org.il/
22. Ali, A., & Renals, S. (2018, July). Word error rate estimation for speech recognition: e-WER. In Proceedings of the 56th Annual Meeting of the Association for Computational Linguistics (Volume 2: Short Papers) (pp. 20-24